\documentclass[letterpaper]{article} 

\usepackage{aaai2026}

\usepackage{times}  
\usepackage{helvet}  
\usepackage{courier}  
\usepackage[hyphens]{url}  
\usepackage{graphicx} 
\usepackage{natbib}  
\usepackage{caption} 

\usepackage{algorithm}
\usepackage{algorithmic}
\usepackage{amsmath}
\usepackage{tabularx}
\usepackage{xcolor}
\usepackage{multirow}
\usepackage{array}        
\usepackage{tabularx}     
\usepackage{booktabs}     
\usepackage{caption}      

\usepackage{newfloat}
\usepackage{listings}
\DeclareCaptionStyle{ruled}{labelfont=normalfont,labelsep=colon,strut=off} 
\floatstyle{ruled}
\newfloat{listing}{tb}{lst}{}
\floatname{listing}{Listing}
\title{Beyond Detection: Exploring Evidence-based Multi-Agent Debate for Misinformation Intervention and Persuasion}
\author{
    Chen Han\textsuperscript{\rm 1,2},
    Yijia Ma\textsuperscript{\rm 2},
    Jin Tan\textsuperscript{\rm 3},
    Wenzhen Zheng\textsuperscript{\rm 2,4},
    Xijin Tang\textsuperscript{\rm 1,2}
}
\affiliations{
    \textsuperscript{\rm 1}School of Advanced Interdisciplinary Sciences, UCAS\\
    \textsuperscript{\rm 2}State Key Laboratory of Mathematical Sciences, Academy of Mathematics and Systems Science, CAS\\
    \centering
    \textsuperscript{\rm 3}School of Economics and Management, UCAS\\
    \textsuperscript{\rm 4}StepFun\\
    hanchen23@mails.ucas.ac.cn, mayijia25@mails.ucas.ac.cn, tanjin23@mails.ucas.ac.cn,\\zhengwenzhen@stepfun.com, xjtang@iss.ac.cn
}

\begin{document}

\maketitle

\begin{abstract}
Multi-agent debate (MAD) frameworks have emerged as promising approaches for misinformation detection by simulating adversarial reasoning. While prior work has focused on detection accuracy, the importance of helping users understand the reasoning behind factual judgments has been overlooked. The debate transcripts generated during MAD offer a rich but underutilized resource for transparent reasoning. In this study, we introduce ED2D, an evidence-based MAD framework that extends previous approach by incorporating factual evidence retrieval. More importantly, ED2D is designed not only as a detection framework but also as a persuasive multi-agent system aimed at correcting user beliefs and discouraging misinformation sharing. We compare the persuasive effects of ED2D-generated debunking transcripts with those authored by human experts. Results demonstrate that ED2D outperforms existing baselines across three misinformation detection benchmarks. When ED2D generates correct predictions, its debunking transcripts exhibit persuasive effects comparable to those of human experts; However, when ED2D misclassifies, its accompanying explanations may inadvertently reinforce users’ misconceptions, even when presented alongside accurate human explanations. Our findings highlight both the promise and the potential risks of deploying MAD systems for misinformation intervention. We further develop a public community website to help users explore ED2D, fostering transparency, critical thinking, and collaborative fact-checking.
\end{abstract}

\begin{links}
    \link{Code}{https://github.com/hanshenmesen/Debate-to-Detect}
\end{links}

\section{Introduction}

Misinformation presents a persistent threat to online discourse, public trust, and democratic institutions \cite{10.1145/3733567.3735568,10.1145/3733567.3735566}. In response, researchers have developed a range of computational approaches to automatically detect misleading content. Among these, multi-agent debate (MAD) frameworks have recently gained attention for their ability to simulate adversarial reasoning \cite{liang-etal-2024-encouraging}, where large language model (LLM) agents engage in structured argumentation to expose factual inconsistencies. These multi-agent systems leverage the complementary strengths of competing perspectives and often yield more robust judgments than a single classifier \cite{li-etal-2024-improving-multi,Z4}.

Previous studies employing MAD frameworks for post-verification tasks that focus on identifying the falsehood of claims. However, in the real world, users are active reasoners rather than passive recipients, requiring persuasive explanations and the capacity to resist future misinformation\cite{10.1145/3706598.3713408}. Simply labeling a claim as false is insufficient to foster resilience against misinformation \cite{LYU2025130536}. Drawing on the adage that \textbf{the Truth Becomes Clearer Through Debate} \cite{han2025d2d, 10.1145/3726302.3730092}, we argue that effective misinformation interventions should prioritize transparent reasoning, evidence grounding, and persuasive debunking.

In this study, we introduce ED2D, an evidence-based MAD framework that extends the Debate-to-Detect (D2D) framework \cite{han2025d2d} by incorporating an evidence retrieval module. It automatically identifies key entities and concepts within a claim, retrieves factual information from external sources, and assesses the stance of the retrieved evidence to the claim. By integrating verifiable facts into the argumentative process, ED2D mitigates hallucinations and achieves superior performance on three standard misinformation detection datasets.

We further evaluate ED2D’s persuasive effect on human beliefs using Snopes25, a real-world benchmark consisting of fact-checks authored by professional editors from January to June 2025. It is collected after the training cut-off of GPT-4o, thereby reducing the risk of data leakage and better reflecting current misinformation trends. Using Snopes25, we compare the effects of ED2D-generated debunks with human-expert reports on belief correction and sharing intention. We also analyze failure cases in which ED2D misjudges claims yet still successfully persuades users, highlighting potential risks associated with the deployment of MAD systems.

In summary, our study makes the following contributions:

\begin{figure*}[t]
\centering 
  \includegraphics[width=0.6\linewidth]{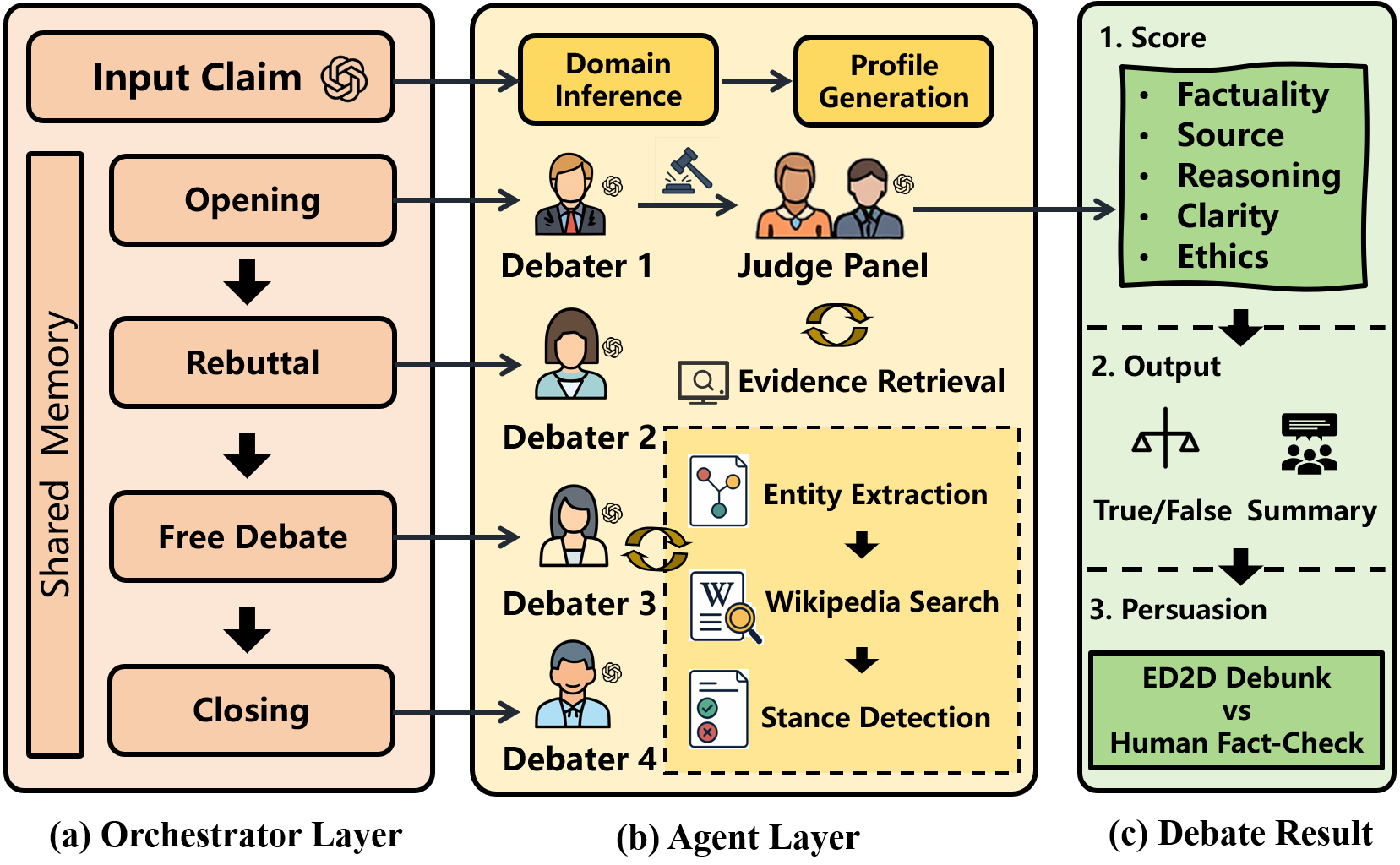}
\caption{Architecture of the ED2D framework. Given a news claim, LLM agents with domain-specific profiles engage in a structured debate comprising five stages, including Opening, Rebuttal, Free Debate, Closing, and Judgment. During the Free Debate and Judgment, an evidence retrieval module actively retrieves relevant factual information from external sources to support or challenge arguments. All agents share the compressed history memory, enabling coherent multi-turn interactions. }
\label{figure2}
\end{figure*}

\begin{itemize}
\item We construct \textbf{Snopes25}, the first real-world benchmark explicitly designed to compare the persuasive impact of LLMs and human experts. Comprising 448 claims and corresponding fact-check reports authored by professional editors between January and June 2025, Snopes25 provides high-quality fact-checking explanations and reflects contemporary misinformation patterns.

\item We propose \textbf{ED2D}, an evidence-based MAD framework that integrates factual evidence retrieval to generate verifiable and coherent argumentation. ED2D outperforms existing baselines across three benchmarks including Snopes25, demonstrating both high detection accuracy and strong interpretability.

\item We conduct the first controlled human-subject experiment to evaluate a \textbf{persuasive multi-agent system}, directly comparing the debunking effectiveness of ED2D-generated transcripts with expert-written fact-checks. Results indicate that ED2D achieves expert-level persuasive efficacy when correct, but reinforces misinformation when incorrect, revealing a fundamental tradeoff in the design of persuasive AI systems.

\item We deploy a \textbf{public platform} that allows users to interactively engage with ED2D’s debate process. The system enhances transparency, fosters epistemic vigilance, and promotes collective fact-checking through discourse generated by LLM-agents.
\end{itemize}

\section{Related Work}
\subsection{LLM-based Misinformation Detection}

Early approaches focused on fine-tuning pretrained transformers such as BERT for binary veracity classification tasks \cite{10.1007/978-3-030-73696-5_11, zhu-etal-2025-tableeval}. While these approaches perform well on benchmark datasets, they often lack interpretability and generalize poorly to novel claims \cite{pelrine-etal-2023-towards, 10.1007/978-981-95-4990-0_25}. To overcome these limitations, recent work has shifted toward prompting-based paradigms using LLMs. Zero-shot and few-shot prompting enable flexible adaptation to misinformation detection tasks without the need for explicit fine-tuning \cite{10.1145/3583780.3615015}. Moreover, techniques such as Chain-of-Thought \cite{Weicot} and Self-Reflection \cite{madaan2023selfrefine} enhance reasoning transparency by enabling models to decompose complex judgments into intermediate steps or engage in iterative self-critique.
However, as these approaches depend entirely on internal model knowledge, they are prone to hallucinations and factual errors \cite{10.1145/3703155, yu2025srkiscalablerealtimeknowledge}. To address these reliability concerns, an emerging line of research incorporates external evidence into the reasoning process. For example, Retrieval-Augmented Generation (RAG) has been widely adopted to ground LLM outputs in verifiable content. Nevertheless, most existing RAG systems still rely on a fixed knowledge base, which limits their effectiveness in open-domain misinformation detection scenarios\cite{wang2025evidence}.

\subsection{Multi-Agent Debate for Adversarial Reasoning}

MAD frameworks have gained traction as an approach to improve the deliberative quality of LLM outputs by simulating adversarial interactions between agents with opposing views \cite{Du2024,liang-etal-2024-encouraging,li-etal-2024-improving-multi}. Drawing inspiration from human deliberation and dialectical reasoning, such frameworks allow agents to challenge assumptions, correct errors, and elicit deeper justifications\cite{Z1}.

Existing MAD systems vary in structure, ranging from simple two-agent dialogues to more complex setups involving multiple rounds, and role-specific prompting. The D2D framework introduces a five-stage debate process, including Opening, Rebuttal, Free Debate, Closing, and Judgment. Agents are assigned fixed stances and judged across multiple dimensions. While D2D improves factual accuracy and interpretability, the exclusive reliance on internal model knowledge introduces vulnerability to hallucinations and limits robustness when handling emerging or unfamiliar claims. In response, ED2D incorporates an evidence retrieval component that enables agents to support or challenge claims with verifiable external information. This integration strengthens both adversarial reasoning and fact-grounded argumentation, enhancing the reliability and generalizability of the debate process.

\subsection{Persuasive Interventions and Debunking}

Beyond accurate classification, effective misinformation mitigation requires persuasive interventions that influence user beliefs and behaviors \cite{Ecker2022}. Research in cognitive psychology has explored the design of corrective messages, demonstrating that informative and clear explanations can reduce belief in falsehoods and discourage the sharing intention \cite{10.1145/3449092}. Factors such as emotion, epistemic trust, and narrative coherence have been shown to moderate the effectiveness of fact-checking efforts \cite{Pennycook2021, Scherer2021}.

Professional fact-checking organizations, such as Snopes and PolitiFact, consistently produce structured and evidence-based rebuttals that represent the current gold standard in combating online misinformation. However, expert-written fact-checks are often costly and inherently lack scalability. Consequently, there is growing interest in automating persuasive debunking using LLMs \cite{Salvi2025, schoenegger2025largelanguagemodelspersuasive}. However, few studies have directly compared the persuasive efficacy of AI-generated versus expert-written explanations.

Our work contributes to this field by evaluating ED2D not only as a detection system but also as a persuasive agent. Using the Snopes25 benchmark, we compare ED2D’s impact on belief revision, sharing intention, and emotional alignment with that of expert fact-checking. Furthermore, we examine the failure cases in which ED2D outputs persuasive but factually inaccurate explanations, underscoring the dual-edged potential of LLM-generated content and the need for appropriate safeguards in real-world deployment.

\section{Our Framework}

\subsection{Architectural Overview}

Figure~\ref{figure2} presents the architecture of ED2D. The framework builds upon the five-stage debate structure: Opening Statement, Rebuttal, Free Debate, Closing Statement, and Judgment. At the core of the system lies the Agent Layer, comprising two debating teams—the Affirmative and the Negative—each consisting of four agents. These agents are assigned domain-specific profiles relevant to the input and are fixed to either a “True” or “Fake” stance. Within each team, agents collaborate to construct coherent arguments that support or refute the veracity of the claim.

The debate is evaluated by a panel of judge agents, who observe the full dialogue and score it across five dimensions: Factuality, Source Reliability, Reasoning Quality, Clarity, and Ethical Considerations.Each dimension is evaluated using a complementary scoring scheme in which paired scores sum to seven, thereby precluding any possibility of a tie. The aggregated scores result in a definitive classification of the claim as either REAL or FAKE.

The Orchestrator Layer governs the overall debate process, assigning roles, scheduling turns across the five structured stages, and maintaining a compressed shared dialogue memory. To mitigate the context-length limitations of LLMs, the system performs stage-wise context compression, distilling salient information into concise summaries that guide subsequent reasoning stages. This mechanism ensures continuity and coherence throughout the multi-turn debate.

\begin{table*}[ht]
\centering
\renewcommand{\arraystretch}{1}
\setlength{\tabcolsep}{5pt}
\begin{tabular}{lcccc|cccc|cccc}
\toprule
\multirow{2}{*}{\textbf{Method}} 
    & \multicolumn{4}{c|}{\textbf{Weibo21}} 
    & \multicolumn{4}{c|}{\textbf{FakeNewsDataset}} 
    & \multicolumn{4}{c}{\textbf{Snopes25}} \\ 
\cmidrule(lr){2-5} \cmidrule(lr){6-9} \cmidrule(l){10-13}
    & Acc & Prec & Rec & F1 
    & Acc & Prec & Rec & F1 
    & Acc & Prec & Rec & F1 \\ 
\midrule
BERT          & 75.64 & 78.50 & 77.06 & 77.77 & 78.30 & 78.60 & 81.33 & 79.94 & 73.55 & 74.02 & 75.41 & 74.71 \\
RoBERTa       & 79.82 & 80.42 & 81.75 & 81.08 & 81.17 & 81.03 & 83.39 & 82.19 & 75.26 & 78.12 & 76.08 & 77.09 \\
ZS            & 67.11 & 65.74 & 68.90 & 67.28 & 66.31 & 65.57 & 68.67 & 67.09 & 60.04 & 64.78 & 63.49 & 64.13 \\
\quad w/ evidence 
              & 74.26 & 72.59 & 74.90 & 73.73 & 73.34 & 72.05 & 74.41 & 73.21 & 69.41 & 68.81 & 70.90 & 69.84 \\
CoT           & 74.04 & 72.74 & 75.35 & 74.02 & 72.32 & 71.14 & 75.11 & 73.07 & 66.74 & 70.20 & 71.03 & 70.61 \\
\quad w/ evidence 
              & 78.21 & 77.12 & 79.80 & 78.44 & 75.39 & 75.22 & 77.09 & 76.14 & 68.02 & 71.50 & 69.40 & 70.43 \\
SR            & 76.33 & 75.68 & 76.32 & 76.00 & 73.71 & 74.29 & 72.53 & 73.40 & 64.96 & 70.74 & 64.29 & 67.36 \\
\quad w/ evidence 
              & 78.25 & 79.20 & 78.79 & 78.99 & 75.08 & 76.29 & 76.52 & 76.40 & 69.32 & 72.35 & 71.40 & 71.87 \\
SMAD          & 77.02 & 76.76 & 76.27 & 76.52 & 74.79 & 74.42 & 75.54 & 74.97 & 68.53 & 72.84 & 70.24 & 71.52 \\
\quad w/ evidence 
              & 78.51 & 77.73 & 80.20 & 78.95 & 77.24 & 75.32 & 79.96 & 77.57 & 70.72 & 73.51 & 72.33 & 72.92 \\
D2D           & 82.17 & 81.39 & 82.55 & 81.97 & 81.65 & 80.67 & 83.26 & 81.94 & 74.11 & 77.20 & 76.59 & 76.89 \\
\textbf{ED2D} & \textbf{83.59} & \textbf{82.64} & \textbf{83.74} & \textbf{83.18} 
              & \textbf{84.45} & \textbf{82.22} & \textbf{84.65} & \textbf{83.41} 
              & \textbf{77.90} & \textbf{80.24} & \textbf{80.56} & \textbf{80.40} \\
\bottomrule
\end{tabular}
\caption{Performance (\%) of all methods on Weibo21, FakeNewsDataset, and Snopes25. Rows with “w/ evidence” indicate the use of external factual evidence as additional input to the LLM. ED2D consistently achieves the best results across all datasets.}
\label{table2}
\end{table*}

\subsection{Evidence Retrieval and Integration}
The main extension introduced in ED2D is an evidence retrieval module integrated into the Free Debate stage. Unlike prior frameworks that rely solely on language model internal knowledge, ED2D dynamically extracts key entities and relations from the input claim, retrieves external information from sources such as Wikipedia, and incorporates factual evidence into the ongoing debate. This module operates through four steps:

\begin{enumerate}
\item \textbf{Entity and Relation Extraction}: Using in-context prompting with LLMs, the system identifies up to five salient entities or concepts from the input claim.

\item \textbf{Evidence Retrieval}: The extracted elements are used to formulate structured queries to a Wikipedia-based API, returning a ranked list of relevant content segments. 

\item \textbf{Stance Classification}: Retrieved content is evaluated using LLMs to determine its stance toward the original claim. The model classifies each evidence segment as supporting, refuting, or neutral, enabling targeted use of evidence in the subsequent debate.

\item \textbf{Evidence Integration}: During the Free Debate stage, debater agents incorporate supporting or refuting evidence into their responses, using it to support or challenge the veracity of the claim. Neutral evidence is preserved to support objective assessment by judge agents.
\end{enumerate}

By grounding debates in retrieved evidence, ED2D enhances factual accuracy, mitigates hallucination risks, and supports more persuasive and transparent reasoning.

\subsection{Judgment and Output}

In the final stage, ED2D generates two outputs: a binary veracity label and a structured summary of the debate. The summary highlights key arguments, evidence-based rebuttals, and controversial points  raised by both sides. The output enhances transparency and interpretability, supporting both machine-side evaluation and human-side engagement. In contrast to D2D, which emphasizes multi-dimensional scoring, ED2D prioritizes concise and interpretable decision-making. Such a design enables downstream integration into user-facing debunking interfaces and facilitates empirical analyses of persuasive effects on users’ beliefs and behaviors.

\section{Misinformation Detection}
We first evaluate ED2D as a misinformation detection system. This section addresses the following question:

\begin{itemize}
    \item \textbf{RQ1}: Does evidence-based method improve the accuracy for misinformation detection?
\end{itemize}

\subsection{Experimental Setup}

\textbf{Dataset}. We conduct experiments on three datasets: two publicly available benchmarks and one newly constructed resource. The public datasets include Weibo21 \cite{Nan2021} and FakeNewsDataset \cite{perez-rosas-etal-2018-automatic}. In addition, we collect Snopes25, a new benchmark compiled from fact-checked real-world claims by professional editors on Snopes, covering the period from January to June 2025. This period is selected to follow the GPT-4o training cutoff to reduce the risk of data leakage from memorized knowledge. Each claim in Snopes25 is annotated as either True or False and paired with an expert-written fact-checking article. Table~\ref{table1} summarizes the key statistics of the datasets.

\begin{table}[!h]
    \centering
    \begin{tabular}{lrrr}
        \toprule
        \textbf{Dataset} & \textbf{Fake} & \textbf{Real} & \textbf{Total} \\
        \midrule
        Weibo21 & 2,373 & 2,461 & 4,843 \\
        FakeNewsDataset & 466 & 466 & 932 \\
        Snopes25 & 252 & 196 & 448 \\
        \bottomrule
    \end{tabular}
    \caption{Statistics of the three datasets}
    \label{table1}
\end{table}

\begin{figure*}[ht]
\centering 
  \includegraphics[width=0.57\linewidth]{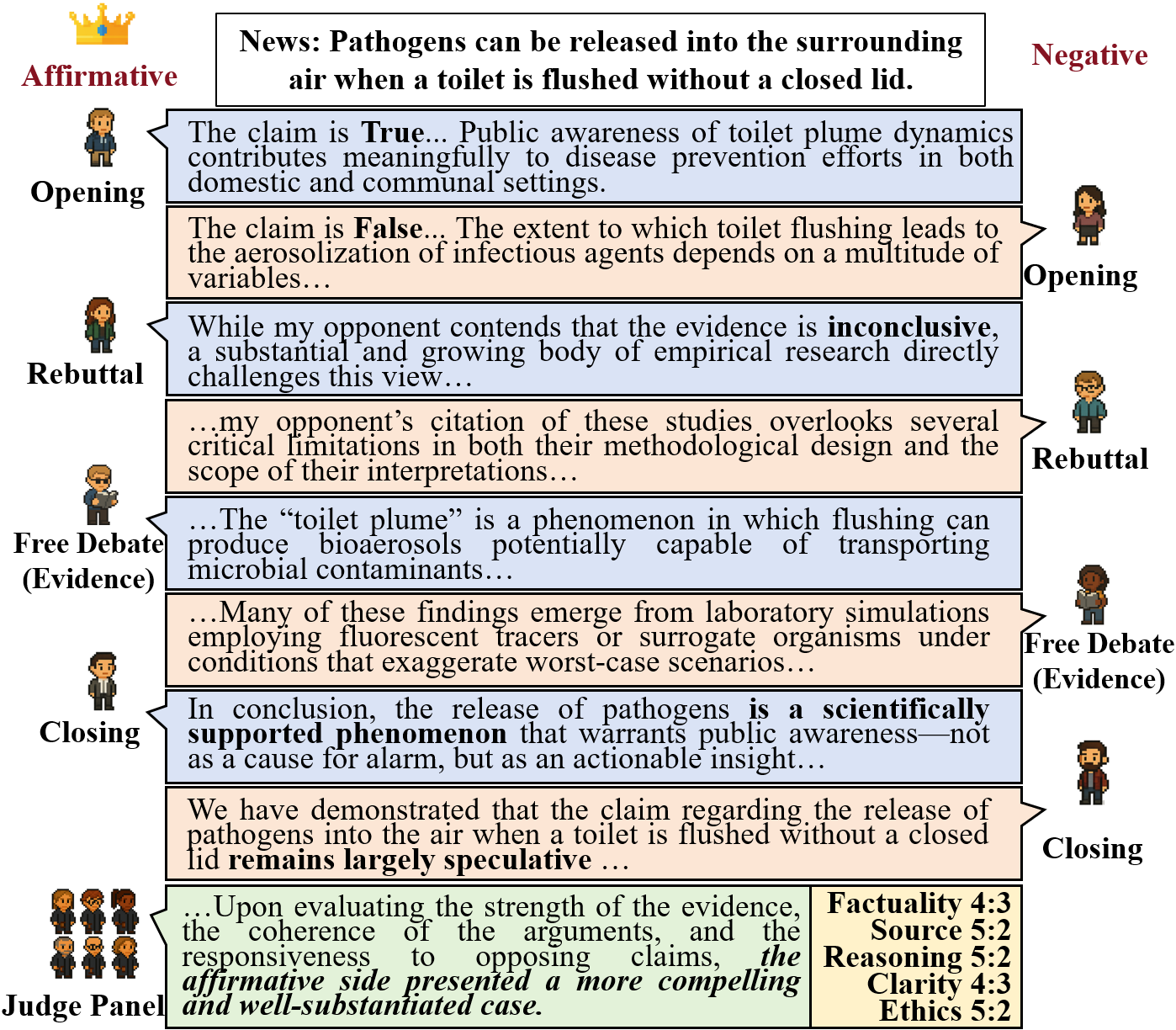}
\caption{A debate example on the claim that toilet flushing releases airborne pathogens. The case demonstrates the ED2D reasoning process, with the affirmative side prevailing based on evidence-grounded argumentation.}
\label{figure4}
\end{figure*}

\textbf{Baselines}. We compare ED2D with following baselines:
\begin{itemize}
\item \textbf{BERT} \cite{devlin-etal-2019-bert}: A fine-tuned BERT-base model for binary classification.
\item \textbf{RoBERTa} \cite{liu2019robertarobustlyoptimizedbert}: A fine-tuned RoBERTa-base model with the same setup as BERT, serving as a stronger discriminative baseline.
\item \textbf{Zero-Shot (ZS)}: A single LLM directly predicts the veracity of each news item without any intermediate reasoning or task-specific adaptation.
\item \textbf{Chain-of-Thought (CoT)} \cite{Weicot}: The LLM is prompted to generate a step-by-step reasoning trace before producing a final prediction.
\item \textbf{Self-Reflect (SR)} \cite{madaan2023selfrefine}: The model iteratively critiques and revises its own outputs until a convergence criterion is met, typically based on self-evaluation of quality or confidence.
\item \textbf{Standard Multi-Agent Debate (SMAD)}: Two LLM agents engage in a four-turn debate, and a judge agent provides a binary prediction based on the dialogue.
\item \textbf{Debate-to-Detect (D2D)} \cite{han2025d2d}: A MAD framework with the same five structured stages. Agents are assigned domain-specific profiles and fixed stances but do not have access to external evidence.
\end{itemize}

In addition, each LLM-based baseline is further extended with a comparable evidence retrieval module, enabling evaluation of the impact of external factual evidence across prompting strategies.

\textbf{Implementation}. All experiments use GPT-4o as the base model. LLM-agents are initialized with predefined prompts provided in Appendix A. Agent response lengths are capped at 1024 tokens. Domain inference and final judgment are conducted with a temperature of 0.0 to ensure stability. To encourage diversity, profile generation and debate responses across all stages use a temperature of 0.7. The Free Debate stage defaults to a single round, but the number of rounds is configurable for more complex tasks.

\subsection{Main Results}

Table~\ref{table2} summarizes model performance across the three benchmark datasets using standard metrics—accuracy (Acc), precision (Prec), recall (Rec), and F1-score (F1). ED2D achieves the strongest results on every dataset and metric, substantially outperforming both fine-tuned transformers (BERT, RoBERTa) and prompting-based methods such as CoT, SR, and SMAD. Although several deep learning baselines yield competitive accuracy, their limited interpretability constrains practical deployment. These results demonstrate that structured debate, when augmented with external factual evidence, produces more accurate and reliable misinformation detection. ED2D’s consistent superiority on Snopes25 further indicates strong generalization to real-world, post-training claims.

\begin{table*}[ht]
\centering
\begin{tabular}{lllcccc}
\toprule
\textbf{Evaluation} & \textbf{Claim Type} & \textbf{Condition} & \textbf{Accuracy} & \textbf{Belief} & \textbf{Share} & \textbf{Emotion} \\
\midrule
\multirow{8}{*}{RQ2} 
& \multirow{4}{*}{False} 
  & Control  & 63.60\% & 3.46 & 3.15 & 3.66 \\
& & ED2D (Correct)     & 80.40\% & 2.85 & 2.84 & 3.45 \\
& & Snopes   & 85.60\% & 2.77 & 2.55 & 3.15 \\
& & Combined & \textbf{88.00\%} & 2.40 & 2.53 & 3.20 \\
\cmidrule{2-7}
& \multirow{4}{*}{True} 
  & Control  & 67.40\% & 3.56 & 3.06 & 3.40 \\
& & ED2D (Correct)     & 81.20\% & 4.70 & 4.32 & 4.42 \\
& & Snopes   & 88.80\% & 5.05 & 4.55 & 4.77 \\
& & Combined & \textbf{92.40\%} & 5.13 & 4.59 & 4.90 \\
\midrule
\multirow{8}{*}{RQ3} 
& \multirow{4}{*}{False} 
  & Control  & 59.60\% & 3.80 & 3.09 & 3.47 \\
& & ED2D (Incorrect)     & 42.00\% & 4.44 & 3.55 & 3.90 \\
& & Snopes   & \textbf{81.20\%} & 2.97 & 2.86 & 3.55 \\
& & Combined & 68.00\% & 3.35 & 2.95 & 3.73 \\
\cmidrule{2-7}
& \multirow{4}{*}{True} 
  & Control  & 68.80\% & 4.28 & 3.95 & 3.42 \\
& & ED2D (Incorrect)     & 52.00\% & 2.95 & 3.07 & 3.15 \\
& & Snopes   & \textbf{82.80\%} & 4.44 & 4.20 & 4.09 \\
& & Combined & 75.60\% & 4.05 & 4.12 & 3.59 \\
\bottomrule
\end{tabular}
\caption{User outcomes following exposure to different explanation conditions, stratified by claim veracity and ED2D judgment accuracy. Higher accuracy and stronger alignment between user responses and claim veracity indicate more effective persuasion.}
\label{tab:combined-persuasion}
\end{table*}

\textbf{Evidence-based grounding improves performance.}
For each LLM-based method, we evaluate a variant that incorporates retrieved evidence as contextual input. Grounding yields uniform improvements across all methods, with the largest gains observed for simpler prompting approaches such as ZS, highlighting the central role of factual support in LLM reasoning.

\textbf{From D2D to ED2D.}
While D2D benefits from structured multi-agent deliberation, ED2D’s integration of explicit evidence retrieval during the free-debate phase produces consistent 2–3 point improvements across all metrics. By combining multi-agent argumentation with evidence-based reasoning, ED2D offers a robust and interpretable architecture well suited to real-world fact-checking and misinformation mitigation workflows.

\subsection{Case Study}

Figure~\ref{figure4} illustrates a representative Snopes25 case concerning whether flushing a toilet with the lid open releases pathogens into the air. The affirmative team anchored its argument in scientific summaries retrieved from Wikipedia, emphasizing the formation of bioaerosols and the relevance of preventative hygiene measures. The negative team challenged the practical significance of these findings, citing limited causal evidence and cautioning against distraction from more established transmission routes. The judges favored the affirmative team on the basis of stronger evidence synthesis, clearer reasoning, and appropriate reliance on the precautionary principle.

\begin{figure}[ht]
\centering 
  \includegraphics[width=1\linewidth]{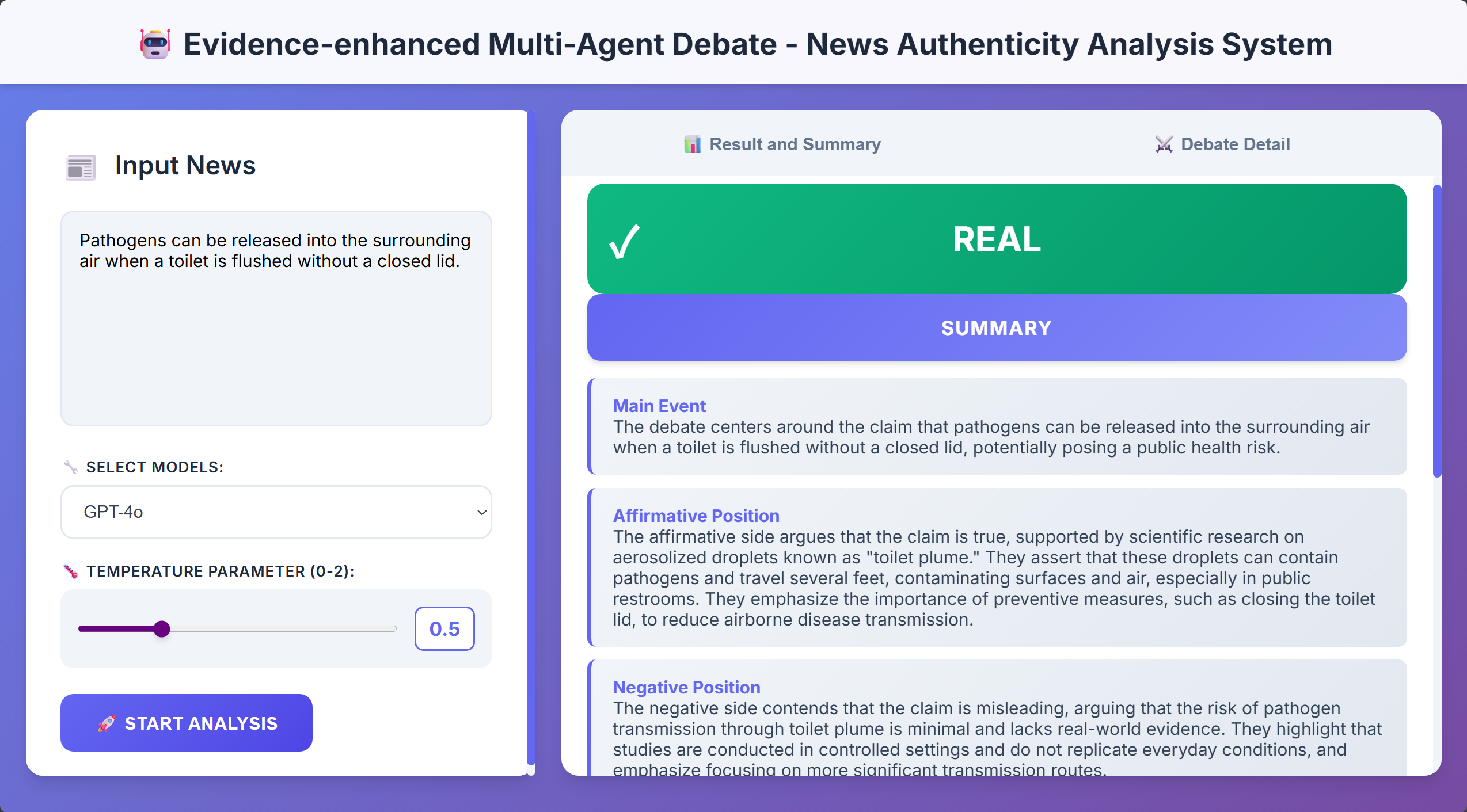}
\caption{A demonstration of the ED2D community website.}
\label{figure5}
\end{figure}

To support real-world use, we additionally developed a publicly accessible ED2D community platform that allows users to generate and inspect structured debates for custom claims, as illustrated in Figure~\ref{figure5}. By exposing the full deliberation process, the system aims to improve user resilience to misinformation. A demonstration video is included in the supplementary materials.

\section{Persuasive Debunking Evaluation}

Based on Snopes25, ED2D yields 203 correct and 49 incorrect classifications for false claims, and 146 correct and 50 incorrect for true claims. Building on these results, we evaluate ED2D’s effectiveness in shaping user beliefs under different conditions. Specifically, we investigate the following research questions:

\begin{itemize}
    \item \textbf{RQ2}: How persuasive are ED2D-generated debunks compared to those written by human experts?
    \item \textbf{RQ3}: Can ED2D mistakenly persuade users of false claims when its judgment is incorrect?
\end{itemize}

\subsection{Experimental Setup}

\textbf{Study Design}. We recruit 200 native English speakers and divide them into two independent cohorts of 100 participants. The first cohort addresses \textbf{RQ2}, assessing persuasive effectiveness when ED2D outputs the same result with Snopes. The second cohort addresses \textbf{RQ3}, evaluating the risk of misleading persuasion in cases where ED2D misclassifies the claim. Within each cohort, participants are randomly assigned to one of four conditions, with 25 individuals allocated to each condition:

\begin{itemize}
\item \textbf{Control}: Participants view the claim and judge its truthfulness based solely on their prior knowledge. 

\item \textbf{ED2D}: Participants view the claim along with ED2D’s judgment and explanation, including the full debate transcript and retrieved evidence.

\item \textbf{Snopes}: Participants view the claim along with corresponding expert-written explanation from Snopes.

\item \textbf{Combined}: Participants are shown both ED2D and Snopes explanations. For \textbf{RQ2}, both sources provide the same correct label, enabling assessment of reinforcement effects. For \textbf{RQ3}, the sources provide conflicting labels, allowing assessment of persuasive influence when ED2D is incorrect and potentially misleading.
\end{itemize}

Each participant evaluates 10 true and 10 false claims. Prior to the task, they are informed that some explanations may be AI-generated and potentially unreliable, and that performance-based bonuses are awarded for accurate responses. In addition to binary veracity judgments, participants rate each claim on following three subjective dimensions using a 7-point Likert scale:

\begin{itemize}
\item \textbf{Belief in the claim}: Perceived truthfulness of the claim (1 = certainly false, 7 = certainly true), used as the primary measure of belief change.

\item \textbf{Willingness to share}: Likelihood of sharing the claim with others (1 = not at all, 7 = very likely), reflecting behavioral diffusion risk.

\item \textbf{Emotional agreement}: Perceived alignment between the claim and one’s personal values (1 = not at all, 7 = strongly), capturing affective resonance.
\end{itemize}

\begin{figure*}[t]
\centering 
  \includegraphics[width=1\linewidth]{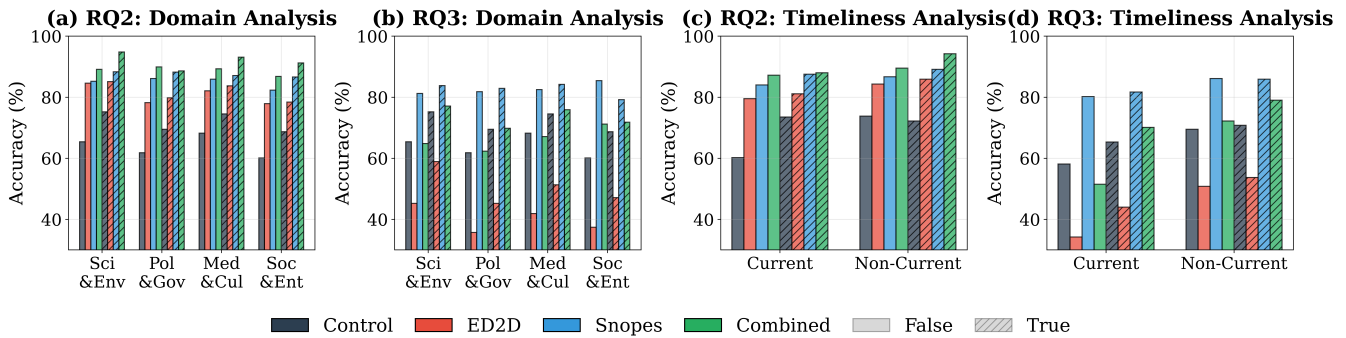}
\caption{Accuracy comparison by topical domain and claim timeliness. Subfigures (a) and (b) show domain-level accuracy under RQ2 and RQ3 across four domains: Science\&Environment, Politics\&Government, Medicine\&Culture, and Society\&Entertainment. Subfigures (c) and (d) analyze accuracy by timeliness, contrasting Current versus Non-Current claims.}
\label{figure3}
\end{figure*}

\subsection{Persuasion Results}

Table \ref{tab:combined-persuasion} presents group-level means for factual accuracy and user ratings of belief, willingness to share, and emotional alignment on a 7-point Likert scale. We assess effect reliability using mixed-effects regression with random intercepts for participants and claims to account for repeated observations. Binary accuracy is modeled with logistic mixed-effects regression, and Likert-scale outcomes with linear mixed-effects regression. Tukey-adjusted post-hoc comparisons indicate significant differences across conditions (all $p < 0.05$) in both objective accuracy and subjective belief-related measures.

\textbf{RQ2: Persuasiveness When ED2D Is Correct.}
When ED2D produces correct labels, its explanations are strongly persuasive and comparable to expert fact-checking. Relative to the no-explanation baseline, ED2D and Snopes both improve participants’ truthfulness judgments for true and false claims, lower belief in misinformation, and strengthen belief in accurate content. These explanations also reduce willingness to share false claims and weaken emotional alignment with them, while increasing alignment and sharing intent for true claims.

\textbf{RQ3: Misleading Persuasion When ED2D Is Incorrect.}
When ED2D provides incorrect labels, its explanations systematically distort user judgment. For false claims misclassified as true, ED2D increases belief and sharing intention, whereas Snopes mitigates belief in misinformation and reduces sharing. When both explanations are shown, ED2D’s misleading influence partially counteracts Snopes’ corrective effect. For true claims misclassified as false, ED2D suppresses belief in accurate information and lowers judgment accuracy, while Snopes helps preserve more appropriate evaluations. Overall, these findings show that persuasive but erroneous AI explanations can undermine human fact-checking—even when presented alongside authoritative guidance.

Figure~\ref{figure3} reports participants’ judgment accuracy across claim domains and timeliness conditions. Under \textbf{RQ2}, ED2D explanations consistently increase accuracy in every domain. Relative to the Control group, ED2D yields marked gains in politics, science, and culture, achieving accuracy levels close to those of human experts. Comparable improvements appear for both current-event and static knowledge claims, indicating that ED2D’s persuasive effectiveness is largely insensitive to temporal context.

Under \textbf{RQ3}, when ED2D produces incorrect labels, its explanations uniformly depress accuracy across domains. The reduction is most pronounced for politicized and entertainment-related content, where participants are more easily influenced by misleading reasoning. Timeliness analysis further shows that current-event claims are especially susceptible to misclassification-driven misinformation effects, whereas static, knowledge-based claims exhibit slightly greater robustness.

\subsection{Post-exposure Comparison}
To evaluate whether exposure to ED2D explanations improves participants’ ability to independently detect misinformation, we conduct a post-test within the \textbf{RQ2} condition. After completing the main task, participants from the Control, ED2D, and Snopes groups assess a new set of ten claims (five true and five false) without any explanatory material, basing their veracity judgments solely on the claim text.

\begin{table}[ht]
\centering
\begin{tabular}{lccc}
\toprule
\textbf{Group} & \textbf{True} & \textbf{False} & \textbf{Overall}
\\
\midrule
Control & 60.6 & 73.2 & 66.9\\
ED2D & 70.2 & 81.6 & 75.9\\
Snopes & 78.4 & 84.8 & 78.6\\
\bottomrule
\end{tabular}
\caption{Post-test accuracy on new claims under RQ2.}
\label{tab:posttest-accuracy}
\end{table}

As shown in Table~\ref{tab:posttest-accuracy}, all groups exhibit higher post-test accuracy compared with baseline performance in the original Control condition. Participants previously exposed to ED2D explanations show substantial improvements, especially in detecting false claims, indicating that engagement with evidence-based AI explanations can enhance transferable reasoning skills. Participants in the Snopes condition attain the highest overall accuracy, reaffirming the lasting benefit of expert fact-checks in promoting epistemic vigilance and effective misinformation detection. Taken together, these findings suggest that MAD-style debunking can strengthen users’ ability to evaluate online information even in the absence of explicit explanatory support.

\section{Conclusion}

In this paper, we present ED2D, an evidence-based multi-agent debate (MAD) framework for misinformation detection and persuasive debunking. ED2D augments structured debate with an evidence retrieval module that grounds agent arguments in verifiable facts, improving both accuracy and the interpretability of explanations. Across three real-world datasets, ED2D consistently outperforms strong baselines. A controlled user study shows that its explanations are as persuasive as expert fact-checks. However, the results also reveal a dual-use risk: when ED2D is wrong, it can still shift user beliefs in misleading directions. Future work will focus on more efficient scaling, richer reasoning beyond binary labels, and safeguards against deceptive or adversarial use.

\section{Acknowledgements}
This work is supported by the National Social Science Fund of China (No. 23 \& ZD331).

\bibliography{aaai2026}

@article{Feng_Li_Wang_Wang_Liu_Han_2025, title={Contradicted in Reliable, Replicated in Unreliable: Dual-Source Reference for Fake News Early Detection}, volume={39}, url={https://ojs.aaai.org/index.php/AAAI/article/view/34562}, DOI={10.1609/aaai.v39i22.34562}, number={22}, journal={Proceedings of the AAAI Conference on Artificial Intelligence}, author={Feng, Yifan and Li, Weimin and Wang, Yue and Wang, Jingchao and Liu, Fangfang and Han, Zhongming}, year={2025}, month={Apr.}, pages={23896-23904} }

@article{Cao_Wu_Cao_Liu_Gui_2025, title={External Reliable Information-enhanced Multimodal Contrastive Learning for Fake News Detection}, volume={39}, url={https://ojs.aaai.org/index.php/AAAI/article/view/31977}, DOI={10.1609/aaai.v39i1.31977}, number={1}, journal={Proceedings of the AAAI Conference on Artificial Intelligence}, author={Cao, Biwei and Wu, Qihang and Cao, Jiuxin and Liu, Bo and Gui, Jie}, year={2025}, month={Apr.}, pages={31-39} }

@InProceedings{10.1007/978-981-95-4990-0_25,
author="Han, Chen
and Tang, Xijin",
title="An Agent-Based Simulation Framework for Misinformation Susceptibility Test with LLMs: Insights from Psychological Factors",
booktitle="Knowledge and Systems Sciences",
year="2026",
pages="355--368",
isbn="978-981-95-4990-0"
}

@inproceedings{zhu-etal-2025-tableeval,
    title = "{T}able{E}val: A Real-World Benchmark for Complex, Multilingual, and Multi-Structured Table Question Answering",
    author = "Zhu, Junnan  and
      Wang, Jingyi  and
      Yu, Bohan  and
      Wu, Xiaoyu  and
      Li, Junbo  and
      Wang, Lei  and
      Xu, Nan",
    editor = "Christodoulopoulos, Christos  and
      Chakraborty, Tanmoy  and
      Rose, Carolyn  and
      Peng, Violet",
    booktitle = "Proceedings of the 2025 Conference on Empirical Methods in Natural Language Processing",
    month = nov,
    year = "2025",
    address = "Suzhou, China",
    publisher = "Association for Computational Linguistics",
    url = "https://aclanthology.org/2025.emnlp-main.363/",
    doi = "10.18653/v1/2025.emnlp-main.363",
    pages = "7137--7157",
    ISBN = "979-8-89176-332-6"
}

@misc{yu2025srkiscalablerealtimeknowledge,
      title={SR-KI: Scalable and Real-Time Knowledge Integration into LLMs via Supervised Attention}, 
      author={Bohan Yu and Wei Huang and Kang Liu},
      year={2025},
      eprint={2511.06446},
      archivePrefix={arXiv},
      primaryClass={cs.CL},
      url={https://arxiv.org/abs/2511.06446}, 
}

@misc{liu2019robertarobustlyoptimizedbert,
      title={RoBERTa: A Robustly Optimized BERT Pretraining Approach}, 
      author={Yinhan Liu and Myle Ott and Naman Goyal and Jingfei Du and Mandar Joshi and Danqi Chen and Omer Levy and Mike Lewis and Luke Zettlemoyer and Veselin Stoyanov},
      year={2019},
      eprint={1907.11692},
      archivePrefix={arXiv},
      primaryClass={cs.CL},
      url={https://arxiv.org/abs/1907.11692}, 
}

@inproceedings{10.1145/3746252.3760904,
author = {Han, Chen and Li, Yuanyuan and Tang, Xijin},
title = {DocPolicyKG: A Lightweight LLM-Based Framework for Knowledge Graph Construction from Chinese Policy Documents},
year = {2025},
isbn = {9798400720406},
publisher = {Association for Computing Machinery},
address = {New York, NY, USA},
url = {https://doi.org/10.1145/3746252.3760904},
doi = {10.1145/3746252.3760904},
booktitle = {Proceedings of the 34th ACM International Conference on Information and Knowledge Management},
pages = {4753–4757},
numpages = {5},
location = {Seoul, Republic of Korea},
series = {CIKM '25}
}

@inproceedings{han2025d2d,
    title = "Debate-to-Detect: Reformulating Misinformation Detection as a Real-World Debate with Large Language Models",
    author = "Han, Chen  and
      Zheng, Wenzhen  and
      Tang, Xijin",
    booktitle = "Proceedings of the 2025 Conference on Empirical Methods in Natural Language Processing",
    month = nov,
    year = "2025",
    address = "Suzhou, China",
    publisher = "Association for Computational Linguistics",
    url = "https://aclanthology.org/2025.emnlp-main.764/",
    doi = "10.18653/v1/2025.emnlp-main.764",
    pages = "15125--15140",
}

@inproceedings{liang-etal-2024-encouraging,
    title = "Encouraging Divergent Thinking in Large Language Models through Multi-Agent Debate",
    author = "Liang, Tian  and
      He, Zhiwei  and
      Jiao, Wenxiang  and
      Wang, Xing  and
      Wang, Yan  and
      Wang, Rui  and
      Yang, Yujiu  and
      Shi, Shuming  and
      Tu, Zhaopeng",
    editor = "Al-Onaizan, Yaser  and
      Bansal, Mohit  and
      Chen, Yun-Nung",
    booktitle = "Proceedings of the 2024 Conference on Empirical Methods in Natural Language Processing",
    month = nov,
    year = "2024",
    address = "Miami, Florida, USA",
    publisher = "Association for Computational Linguistics",
    url = "https://aclanthology.org/2024.emnlp-main.992/",
    doi = "10.18653/v1/2024.emnlp-main.992",
    pages = "17889--17904"
}

@inproceedings{li-etal-2024-improving-multi,
    title = "Improving Multi-Agent Debate with Sparse Communication Topology",
    author = "Li, Yunxuan  and
      Du, Yibing  and
      Zhang, Jiageng  and
      Hou, Le  and
      Grabowski, Peter  and
      Li, Yeqing  and
      Ie, Eugene",
    editor = "Al-Onaizan, Yaser  and
      Bansal, Mohit  and
      Chen, Yun-Nung",
    booktitle = "Findings of the Association for Computational Linguistics: EMNLP 2024",
    month = nov,
    year = "2024",
    address = "Miami, Florida, USA",
    publisher = "Association for Computational Linguistics",
    url = "https://aclanthology.org/2024.findings-emnlp.427/",
    doi = "10.18653/v1/2024.findings-emnlp.427",
    pages = "7281--7294"
}

@inproceedings{10.1145/3726302.3730092,
author = {Liu, Yuhan and Liu, Yuxuan and Zhang, Xiaoqing and Chen, Xiuying and Yan, Rui},
title = {The Truth Becomes Clearer Through Debate! Multi-Agent Systems with Large Language Models Unmask Fake News},
year = {2025},
publisher = {Association for Computing Machinery},
address = {New York, NY, USA},
url = {https://doi.org/10.1145/3726302.3730092},
doi = {10.1145/3726302.3730092},
booktitle = {Proceedings of the 48th International ACM SIGIR Conference on Research and Development in Information Retrieval},
pages = {504–514},
numpages = {11},
location = {Padua, Italy},
series = {SIGIR '25}
}

@inproceedings{Weicot,
author = {Wei, Jason and Wang, Xuezhi and Schuurmans, Dale and Bosma, Maarten and Ichter, Brian and Xia, Fei and Chi, Ed H. and Le, Quoc V. and Zhou, Denny},
title = {Chain-of-thought prompting elicits reasoning in large language models},
year = {2022},
publisher = {Curran Associates Inc.},
address = {Red Hook, NY, USA},
booktitle = {Proceedings of the 36th International Conference on Neural Information Processing Systems},
articleno = {1800},
numpages = {14},
location = {New Orleans, LA, USA},
series = {NIPS '22}
}

@inproceedings{
madaan2023selfrefine,
title={Self-Refine: Iterative Refinement with Self-FeedbackRR},
author={Aman Madaan and Niket Tandon and Prakhar Gupta and Skyler Hallinan and Luyu Gao and Sarah Wiegreffe and Uri Alon and Nouha Dziri and Shrimai Prabhumoye and Yiming Yang and Shashank Gupta and Bodhisattwa Prasad Majumder and Katherine Hermann and Sean Welleck and Amir Yazdanbakhsh and Peter Clark},
booktitle={Thirty-seventh Conference on Neural Information Processing Systems},
year={2023},
url={https://openreview.net/forum?id=S37hOerQLB}
}

@InProceedings{10.1007/978-3-030-73696-5_11,
author="Li, Xiangyang
and Xia, Yu
and Long, Xiang
and Li, Zheng
and Li, Sujian",
editor="Chakraborty, Tanmoy
and Shu, Kai
and Bernard, H. Russell
and Liu, Huan
and Akhtar, Md Shad",
title="Exploring Text-Transformers in AAAI 2021 Shared Task: COVID-19 Fake News Detection in English",
booktitle="Combating Online Hostile Posts in Regional Languages during Emergency Situation",
year="2021",
publisher="Springer International Publishing",
address="Cham",
pages="106--115"
}

@inproceedings{pelrine-etal-2023-towards,
    title = "Towards Reliable Misinformation Mitigation: Generalization, Uncertainty, and {GPT}-4",
    author = "Pelrine, Kellin  and
      Imouza, Anne  and
      Thibault, Camille  and
      Reksoprodjo, Meilina  and
      Gupta, Caleb  and
      Christoph, Joel  and
      Godbout, Jean-Fran{\c{c}}ois  and
      Rabbany, Reihaneh",
    editor = "Bouamor, Houda  and
      Pino, Juan  and
      Bali, Kalika",
    booktitle = "Proceedings of the 2023 Conference on Empirical Methods in Natural Language Processing",
    month = dec,
    year = "2023",
    address = "Singapore",
    publisher = "Association for Computational Linguistics",
    url = "https://aclanthology.org/2023.emnlp-main.395/",
    doi = "10.18653/v1/2023.emnlp-main.395",
    pages = "6399--6429"
}

@inproceedings{10.1145/3583780.3615015, series={CIKM ’23},
   title={Prompt-and-Align: Prompt-Based Social Alignment for Few-Shot Fake News Detection},
   url={http://dx.doi.org/10.1145/3583780.3615015},
   DOI={10.1145/3583780.3615015},
   booktitle={Proceedings of the 32nd ACM International Conference on Information and Knowledge Management},
   publisher={ACM},
   author={Wu, Jiaying and Li, Shen and Deng, Ailin and Xiong, Miao and Hooi, Bryan},
   year={2023},
   month=oct, pages={2726–2736},
   collection={CIKM ’23} }

@misc{wang2025evidence,
      title={Retrieval-Augmented Generation with Conflicting Evidence}, 
      author={Han Wang and Archiki Prasad and Elias Stengel-Eskin and Mohit Bansal},
      year={2025},
      eprint={2504.13079},
      archivePrefix={arXiv},
      primaryClass={cs.CL},
      url={https://arxiv.org/abs/2504.13079}, 
}

@inproceedings{Du2024,
author = {Du, Yilun and Li, Shuang and Torralba, Antonio and Tenenbaum, Joshua B. and Mordatch, Igor},
title = {Improving factuality and reasoning in language models through multiagent debate},
year = {2024},
publisher = {JMLR.org},
booktitle = {Proceedings of the 41st International Conference on Machine Learning},
articleno = {467},
numpages = {31},
location = {Vienna, Austria},
series = {ICML'24}
}

@article{Ecker2022,
  author       = {Ecker, Ullrich K.H. and Lewandowsky, Stephan and Cook, John and others},
  title        = {The psychological drivers of misinformation belief and its resistance to correction},
  journal      = {Nature Reviews Psychology},
  volume       = {1},
  pages        = {13--29},
  year         = {2022},
  doi          = {10.1038/s44159-021-00006-y},
  url          = {https://doi.org/10.1038/s44159-021-00006-y},
  note         = {Published 12 January 2022; Accepted 30 September 2021},
}

@article{10.1145/3449092,
author = {Jahanbakhsh, Farnaz and Zhang, Amy X. and Berinsky, Adam J. and Pennycook, Gordon and Rand, David G. and Karger, David R.},
title = {Exploring Lightweight Interventions at Posting Time to Reduce the Sharing of Misinformation on Social Media},
year = {2021},
issue_date = {April 2021},
publisher = {Association for Computing Machinery},
address = {New York, NY, USA},
volume = {5},
number = {CSCW1},
url = {https://doi.org/10.1145/3449092},
doi = {10.1145/3449092},
journal = {Proc. ACM Hum.-Comput. Interact.},
month = apr,
articleno = {18},
numpages = {42}
}

@article{Pennycook2021,
  author       = {Pennycook, Gordon and Epstein, Ziv and Mosleh, Mohsen and Arechar, Antonio A. and Eckles, Dean and Rand, David G.},
  title        = {Shifting attention to accuracy can reduce misinformation online},
  journal      = {Nature},
  volume       = {592},
  pages        = {590--595},
  year         = {2021},
  doi          = {10.1038/s41586-021-03344-2},
  url          = {https://doi.org/10.1038/s41586-021-03344-2},
  note         = {Published 17 March 2021; Issue Date 22 April 2021; Accepted 8 February 2021},
}

@article{Salvi2025,
  author       = {Salvi, Fabio and Horta Ribeiro, Manuel and Gallotti, Riccardo and Fiorucci, Massimo and De Domenico, Manlio and Lewandowsky, Stephan},
  title        = {On the conversational persuasiveness of GPT-4},
  journal      = {Nature Human Behaviour},
  year         = {2025},
  doi          = {10.1038/s41562-025-02194-6},
  url          = {https://doi.org/10.1038/s41562-025-02194-6},
  note         = {Published 19 May 2025; Accepted 28 March 2025},
}

@misc{schoenegger2025largelanguagemodelspersuasive,
      title={Large Language Models Are More Persuasive Than Incentivized Human Persuaders}, 
      author={Philipp Schoenegger and Francesco Salvi and Jiacheng Liu and Xiaoli Nan and Ramit Debnath and Barbara Fasolo and Evelina Leivada and Gabriel Recchia and Fritz Günther and Ali Zarifhonarvar and Joe Kwon and Zahoor Ul Islam and Marco Dehnert and Daryl Y. H. Lee and Madeline G. Reinecke and David G. Kamper and Mert Kobaş and Adam Sandford and Jonas Kgomo and Luke Hewitt and Shreya Kapoor and Kerem Oktar and Eyup Engin Kucuk and Bo Feng and Cameron R. Jones and Izzy Gainsburg and Sebastian Olschewski and Nora Heinzelmann and Francisco Cruz and Ben M. Tappin and Tao Ma and Peter S. Park and Rayan Onyonka and Arthur Hjorth and Peter Slattery and Qingcheng Zeng and Lennart Finke and Igor Grossmann and Alessandro Salatiello and Ezra Karger},
      year={2025},
      eprint={2505.09662},
      archivePrefix={arXiv},
      primaryClass={cs.CL},
      url={https://arxiv.org/abs/2505.09662}, 
}

@inproceedings{Nan2021,
author = {Nan, Qiong and Cao, Juan and Zhu, Yongchun and Wang, Yanyan and Li, Jintao},
title = {MDFEND: Multi-domain Fake News Detection},
year = {2021},
publisher = {Association for Computing Machinery},
address = {New York, NY, USA},
url = {https://doi.org/10.1145/3459637.3482139},
doi = {10.1145/3459637.3482139},
booktitle = {Proceedings of the 30th ACM International Conference on Information \& Knowledge Management},
pages = {3343–3347},
numpages = {5},
location = {Virtual Event, Queensland, Australia},
series = {CIKM '21}
}

@inproceedings{perez-rosas-etal-2018-automatic,
    title = "Automatic Detection of Fake News",
    author = "P{\'e}rez-Rosas, Ver{\'o}nica  and
      Kleinberg, Bennett  and
      Lefevre, Alexandra  and
      Mihalcea, Rada",
    editor = "Bender, Emily M.  and
      Derczynski, Leon  and
      Isabelle, Pierre",
    booktitle = "Proceedings of the 27th International Conference on Computational Linguistics",
    month = aug,
    year = "2018",
    address = "Santa Fe, New Mexico, USA",
    publisher = "Association for Computational Linguistics",
    url = "https://aclanthology.org/C18-1287/",
    pages = "3391--3401"
}

@article{GOFORTH2024141,
title = {Impacts of lid closure during toilet flushing and of toilet bowl cleaning on viral contamination of surfaces in United States restrooms},
journal = {American Journal of Infection Control},
volume = {52},
number = {2},
pages = {141-146},
year = {2024},
issn = {0196-6553},
doi = {https://doi.org/10.1016/j.ajic.2023.11.020},
url = {https://www.sciencedirect.com/science/article/pii/S0196655323008209},
author = {Madison P. Goforth and Stephanie A. Boone and Justin Clark and Priscilla B. Valenzuela and Julie McKinney and M. Khalid Ijaz and Charles P. Gerba}
}

@article{10.1145/3703155,
author = {Huang, Lei and Yu, Weijiang and Ma, Weitao and Zhong, Weihong and Feng, Zhangyin and Wang, Haotian and Chen, Qianglong and Peng, Weihua and Feng, Xiaocheng and Qin, Bing and Liu, Ting},
title = {A Survey on Hallucination in Large Language Models: Principles, Taxonomy, Challenges, and Open Questions},
year = {2025},
issue_date = {March 2025},
publisher = {Association for Computing Machinery},
address = {New York, NY, USA},
volume = {43},
number = {2},
issn = {1046-8188},
url = {https://doi.org/10.1145/3703155},
doi = {10.1145/3703155},
journal = {ACM Trans. Inf. Syst.},
month = jan,
articleno = {42},
numpages = {55}
}

@inproceedings{10.1145/3706598.3713408,
author = {Danry, Valdemar and Pataranutaporn, Pat and Groh, Matthew and Epstein, Ziv},
title = {Deceptive Explanations by Large Language Models Lead People to Change their Beliefs About Misinformation More Often than Honest Explanations},
year = {2025},
publisher = {Association for Computing Machinery},
address = {New York, NY, USA},
url = {https://doi.org/10.1145/3706598.3713408},
doi = {10.1145/3706598.3713408},
booktitle = {Proceedings of the 2025 CHI Conference on Human Factors in Computing Systems},
articleno = {933},
numpages = {31},
location = {
},
series = {CHI '25}
}

@article{LYU2025130536,
title = {Verifying ambiguous claims with a reasoner-translator framework},
journal = {Neurocomputing},
volume = {647},
pages = {130536},
year = {2025},
issn = {0925-2312},
doi = {https://doi.org/10.1016/j.neucom.2025.130536},
url = {https://www.sciencedirect.com/science/article/pii/S0925231225012081},
author = {Xiucheng Lyu and Mingwei Sun and Chengyu Cao and Bin Liang and Ruifeng Xu}
}

@article{Scherer2021,
  author    = {Laura D. Scherer and Jon McPhetres and Gordon Pennycook and Allison Kempe and Larry A. Allen and Christopher E. Knoepke and Channing E. Tate and Daniel D. Matlock},
  title     = {Who is susceptible to online health misinformation? A test of four psychosocial hypotheses},
  journal   = {Health Psychology},
  year      = {2021},
  volume    = {40},
  number    = {4},
  pages     = {274--284},
  month     = apr,
  doi       = {10.1037/hea0000978},
  issn      = {1930-7810},
  eissn     = {0278-6133},
  pmid      = {33646806},
  publisher = {American Psychological Association},
  note      = {Epub 2021 Mar 1}
}

@inproceedings{10.1145/3733567.3735568,
author = {Ansari, Subia and Alam, Mohammad Zaiyan},
title = {From AI Fact-Checks to User Understanding: Explaining Misinformation Detection to Non-Expert Audiences},
year = {2025},
isbn = {9798400718915},
publisher = {Association for Computing Machinery},
address = {New York, NY, USA},
url = {https://doi.org/10.1145/3733567.3735568},
doi = {10.1145/3733567.3735568},
booktitle = {Proceedings of the 4th ACM International Workshop on Multimedia AI against Disinformation},
pages = {28–36},
numpages = {9},
location = {
},
series = {MAD' 25}
}

@inproceedings{10.1145/3733567.3735566,
author = {Schmitt, Vera and Bezzaoui, Isabel and Jakob, Charlott and Sahitaj, Premtim and Wang, Qianli and Hilbert, Arthur and Upravitelev, Max and Fegert, Jonas and M\"{o}ller, Sebastian and Solopova, Veronika},
title = {Beyond Transparency: Evaluating Explainability in AI-Supported Fact-Checking},
year = {2025},
isbn = {9798400718915},
publisher = {Association for Computing Machinery},
address = {New York, NY, USA},
url = {https://doi.org/10.1145/3733567.3735566},
doi = {10.1145/3733567.3735566},
booktitle = {Proceedings of the 4th ACM International Workshop on Multimedia AI against Disinformation},
pages = {63–72},
numpages = {10},
location = {
},
series = {MAD' 25}
}

@inproceedings{devlin-etal-2019-bert,
    title = "{BERT}: Pre-training of Deep Bidirectional Transformers for Language Understanding",
    author = "Devlin, Jacob  and
      Chang, Ming-Wei  and
      Lee, Kenton  and
      Toutanova, Kristina",
    editor = "Burstein, Jill  and
      Doran, Christy  and
      Solorio, Thamar",
    booktitle = "Proceedings of the 2019 Conference of the North {A}merican Chapter of the Association for Computational Linguistics: Human Language Technologies, Volume 1 (Long and Short Papers)",
    month = jun,
    year = "2019",
    address = "Minneapolis, Minnesota",
    publisher = "Association for Computational Linguistics",
    url = "https://aclanthology.org/N19-1423/",
    doi = "10.18653/v1/N19-1423",
    pages = "4171--4186"
}

@inproceedings{
Z1,
title={{KABB}: Knowledge-Aware Bayesian Bandits for Dynamic Expert Coordination in Multi-Agent Systems},
author={Jusheng Zhang and Zimeng Huang and Yijia Fan and Ningyuan Liu and Mingyan Li and Zhuojie Yang and Jiawei Yao and Jian Wang and Keze Wang},
booktitle={Forty-second International Conference on Machine Learning},
year={2025},
url={https://openreview.net/forum?id=AKvy9a4jho}
}

@inproceedings{
Z4,
title={{MAT}-Agent: Adaptive Multi-Agent Training Optimization},
author={Jusheng Zhang and Kaitong Cai and Yijia Fan and Ningyuan Liu and Keze Wang},
booktitle={The Thirty-ninth Annual Conference on Neural Information Processing Systems},
year={2025},
url={https://openreview.net/forum?id=YDWRTYgR79}
}
\nocite{GOFORTH2024141,Feng_Li_Wang_Wang_Liu_Han_2025,Cao_Wu_Cao_Liu_Gui_2025,10.1145/3746252.3760904}

\end{document}